\documentclass[10pt,twocolumn,letterpaper]{article}

\usepackage{cvpr}
\usepackage{times}
\usepackage{epsfig}
\usepackage{graphicx}
\usepackage{amsmath}
\usepackage{amsfonts}
\usepackage{amssymb}
\usepackage{multirow}
\usepackage{array}
\usepackage[square,sort,comma,numbers]{natbib}
\usepackage{graphicx}
\cvprfinalcopy % *** Uncomment this line for the final submission

\ifcvprfinal\fi
\usepackage[breaklinks=true,bookmarks=false]{hyperref}
\begin{document}

%%%%%%%%% TITLE
\title{Qiniu Submission to ActivityNet Challenge 2018}

\author{Xiaoteng Zhang, Yixin Bao, Feiyun Zhang, Kai Hu, Yicheng Wang,\\ Liang Zhu, Qinzhu He, Yining Lin, Jie Shao and Yao Peng\\
Qiniu AtLab\\
Shanghai, China\\
{\tt\small \{zhangxiaoteng, baoyixin, zhangfeiyun, hukai, wangyicheng\}@qiniu.com}\\
{\tt\small \{zhuliang, heqinzhu, linyining, shaojie, pengyao\}@qiniu.com}
}

\maketitle

\begin{abstract}
In this paper, we introduce our submissions for the tasks of trimmed activity recognition (Kinetics)\cite{kay2017kinetics} and trimmed event recognition (Moments in Time)\cite{monfortmoments} for Activitynet Challenge 2018. In the two tasks, non-local neural networks and temporal segment networks are implemented as our base models. Multi-modal cues such as RGB image, optical flow and acoustic signal have also been used in our method. We also propose new non-local-based models for further improvement on the recognition accuracy. The final submissions after ensembling the models achieve 83.5\% top-1 accuracy and 96.8\% top-5 accuracy on the Kinetics validation set, 35.81\% top-1 accuracy and 62.59\% top-5 accuracy on the MIT validation set.
\end{abstract}

\section{Introduction}
Activity Recognition in videos has drawn increasing attention from the research community in recent years. The state-of-the-art benchmark datasets such as ActivityNet, Kinetics, Moments in Times have contributed to the progress in video understanding.

In Activitynet Challenge 2018, we mainly focused on two trimmed video recognition tasks based on Kinetics and Moments in Times datasets respectively. The Kinetic dataset consists of approximately 500,000 video clips, and covers 600 human action classes. Each clip lasts around 10 seconds and is labeled with a single class. Similarly, the Moments in Times dataset is also a trimmed dataset, including a collection of 339 classes of one million labeled 3 second videos. The videos not only involve people, but also describe animals, objects or natural phenomena, which are more complex and ambiguous than the videos in Kinetics.

To recognize actions and events in videos, recent approaches based on deep convolution neural networks have achieved state-of-the-art results. To address the challenge, our solution follows the strategy of non-local neural network and temporal segment network. Particularly, we learn models with multi-modality information of the videos, including RGB, optical flow and audio. We find that these models are complementary with each other. Our final result is an ensemble of these models, and achieves 83.5\% top-1 accuracy and 96.8\% top-5 accuracy on the Kinetics validation set, 35.81\% top-1 accuracy and 62.59\% top-5 accuracy on the MIT validation set.

\section{Our Methods}
\subsection{Temporal Segment Networks}
One of our base model is temporal segment network (TSN)\cite{wang2017temporal}. TSN models long-term temporal information by evenly sampling fixed number of clips from the entire videos. Each sampled clips contain one or several frames / flow stacks, and produce the prediction separately. The video-level prediction is given by the averaged softmax scores of all clips.

We experiment with several state-of-the-art network architectures, such as ResNet, ResNeXt, Inception\cite{szegedy2017inception}, SENet\cite{hu2017squeeze}, DPN\cite{chen2017dual}. These models are pretrained on ImageNet, and have good initial weights for further training. Table \ref{1} and \ref{2} show our TSN results on Kinetics and Moments in Times dataset.
\begin{table}[h]
   \centering
   \begin{tabular}{p{3.5cm}<{\centering}|p{2cm}<{\centering}|p{2cm}<{\centering}}
      \hline
      Models      & Top-1 acc(RGB) & Top-1 acc(Flow)   \\ \hline
      DPN107      & 75.95               & 69.60                  \\ \hline
      ResNext101      & 75.43               & None                  \\ \hline
      SE-ResNet152      & 73.88              & None                  \\ \hline
      InceptionV4     & 73.51               & 68.76                 \\ \hline
      ResNet152      & 72.04              & 67.13                  \\ \hline
      InceptionV3     & 68.52               & 64.08                 \\ \hline
   \end{tabular}
   \caption{Performance of TSN on Kinetics}
   \label{1}
\end{table}
\begin{table}[h]
   \centering
   \begin{tabular}{p{3.5cm}<{\centering}|p{2cm}<{\centering}|p{2cm}<{\centering}}
      \hline
      Models      & Top-1 acc(RGB) & Top-1 acc(Flow)   \\ \hline
      DPN107      & 31.06               & None                \\ \hline
      ResNet152      & 30.21              & None                  \\ \hline
      ResNet269      & None            & $18.53^*$              \\ \hline
      ResNet101      & None           & 22.82                \\ \hline
   \end{tabular}
   \caption{Performance of TSN on Moments in Times dataset.   $\qquad$*: Due to time limit, the training of these models was not finished.}
   \label{2}
\end{table}

\subsection{Acoustic Model}
While most motions can be recognized from visual information, sound contains information in another dimension. We use audio channels as complementary information to visual information to recognize certain classes, especially for the actions with better distinguishability on sound, whistling and barking for example.

We use the raw audio as input into the pre-trained VGGish model\cite{hershey2017cnn}\cite{gemmeke2017audio}, and extract $n\times 128$ dimension features to do classification (n is the length of audio). Besides, we extract MFCC features from raw audio and train with SE-ResNet-50 (Squeeze-and-Excitation Network) and ResNet-50 (Deep residual network\cite{he2016deep}). After ensembling with visual models, we achieve 0.7\% improvement in top 1 error rate.

\begin{figure*}[h]
\centering\includegraphics[width=13cm]{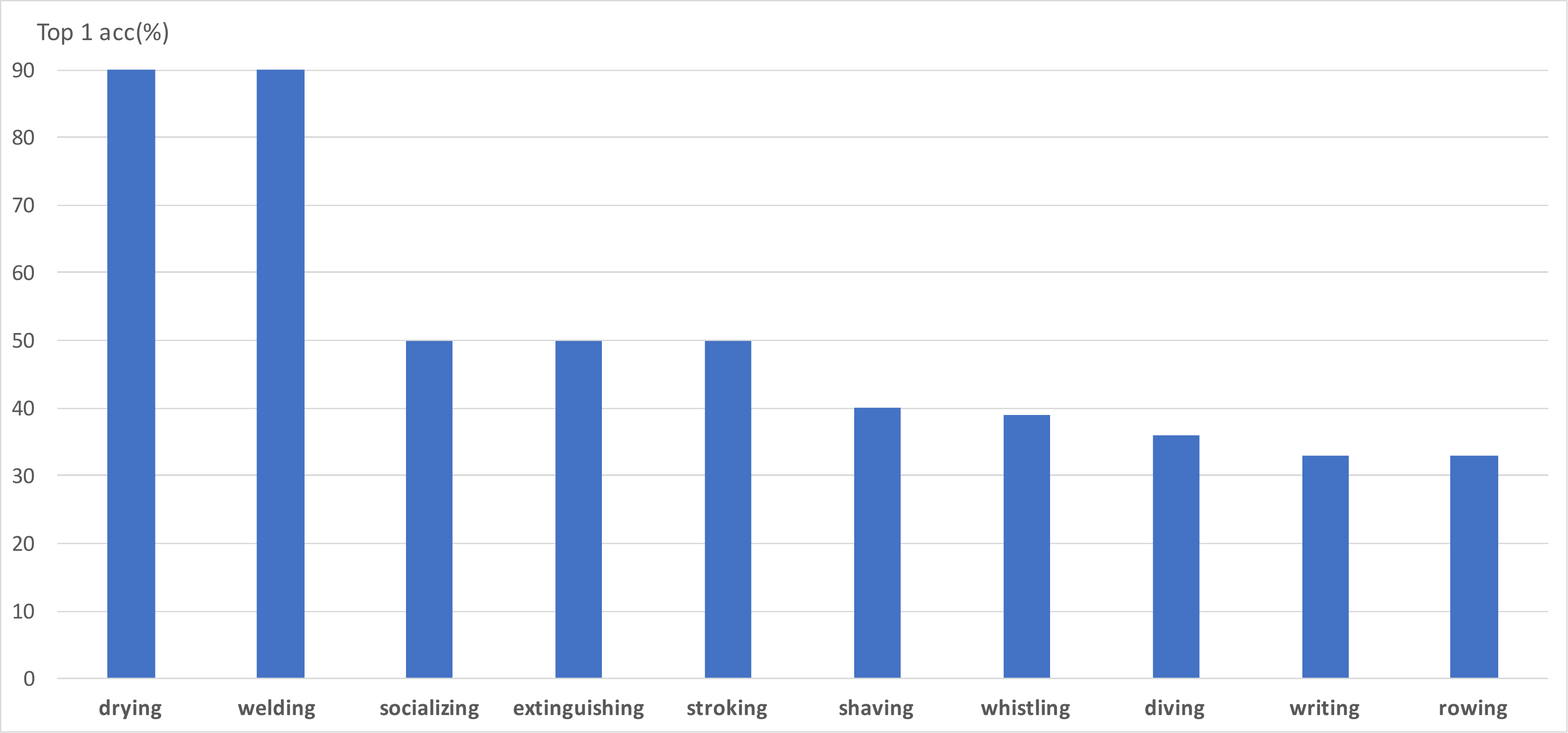}
\caption{Acoustic Model: 10 classes with best top 1 accuracy in MIT validation dataset}\label{78}
\end{figure*}

\begin{figure*}[h]
\centering\includegraphics[width=13cm]{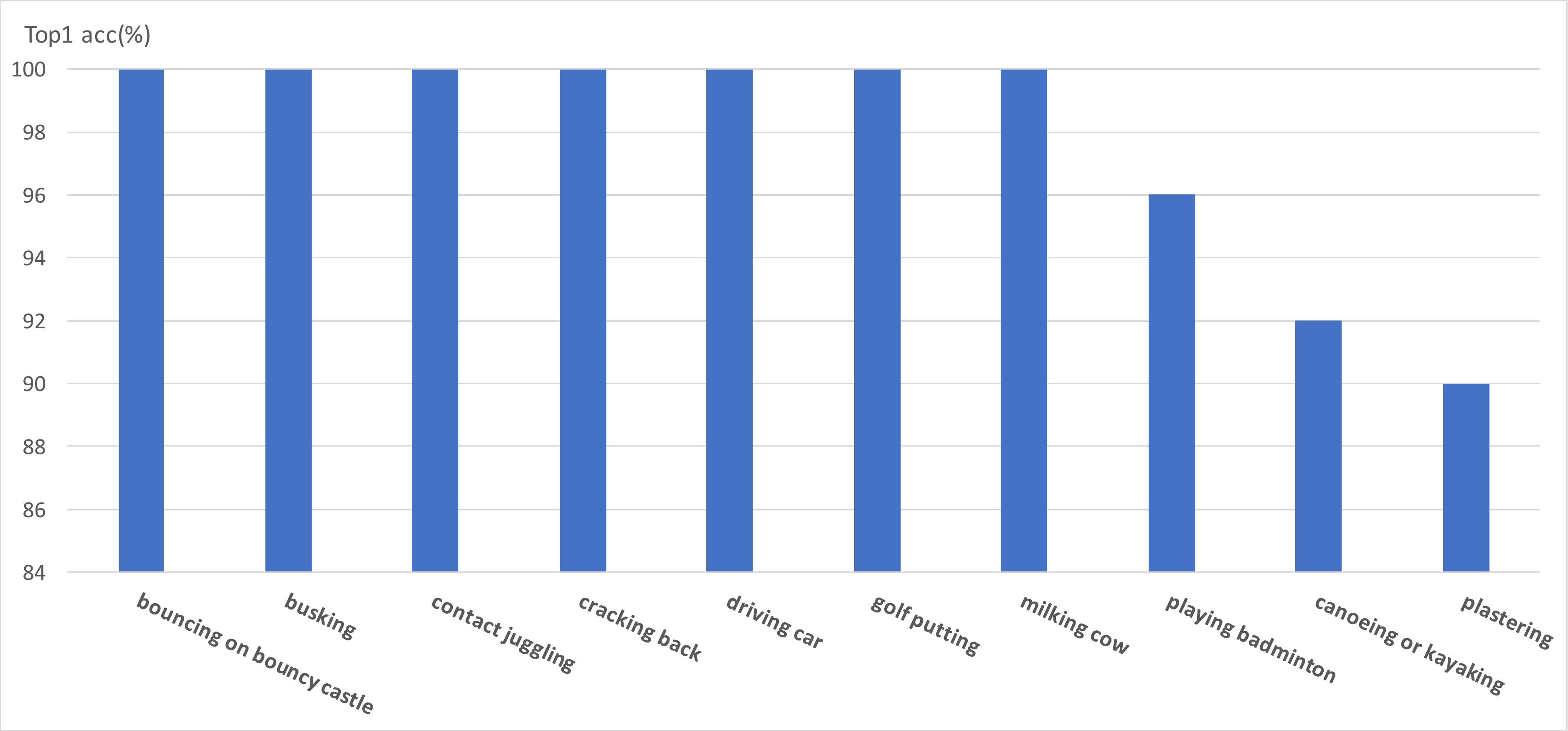}
\caption{Acoustic Model: 10 classes with best top 1 accuracy in Kinetics validation dataset.}\label{56}
\end{figure*}

Table \ref{55555} shows our acoustic results on on Kinetics and Moments in Times dataset. Figure \ref{78} shows 10 classes with best top 1 accuracy in MIT validation dataset and figure \ref{56} shows 10 classes with best top 1 accuracy in Kinetics validation dataset.
\begin{table}[h]
   \centering
   \begin{tabular}{p{3.5cm}<{\centering}|p{2cm}<{\centering}|p{2cm}<{\centering}}
      \hline
      Models      & Kinetics & MIT   \\ \hline
      MFCC+ SENet-50      & 7.73              & 16.8               \\ \hline
      VGGish      & 7.83             & 17.12                 \\ \hline
      Audio Ensemble     & 8.83              & 19.02              \\ \hline
   \end{tabular}
   \caption{Performance of acoustic models on Kinetics and MIT dataset.}
   \label{55555}
\end{table}
\subsection{Non-local Neural Networks}
Non-local Neural Networks\cite{NonLocal2018}  extract long-term temporal information which have demonstrated the  significance of non-local modeling for the tasks of video classification, object detection and so on.

{\bfseries Notation} Image data and feature map data are generally three-dimensional: channel, height and width (in practice there is one more dimension: batch).  They are represented as C-dimensional vectors with 2-dimensional index：
\begin{equation}
\boldsymbol{X} = \{\boldsymbol{x}_i|i=(h,w)\in  \mathbb{D}^2, \boldsymbol{x}_i \in \mathbb{R}^C\}
\end{equation}
where $\mathbb{D}^2=\{1,2,\cdots,H\}\times\{1,2,\cdots,W\}$. Video data get one more dimension time and are represented as C-dimensional vectors with 3-dimensional index：
\begin{equation}
\boldsymbol{X} = \{\boldsymbol{x}_i|i=(t,h,w)\in \mathbb{D}^3, \boldsymbol{x}_i \in \mathbb{R}^C\}
\end{equation}
where $\mathbb{D}^3=\{1,2,\cdots,T\}\times \mathbb{D}^2$.

%%%%%%%%%%%%%%%%%%%%%%%%%%%%%%%%%%%%%%%%%%%
%%%%%%%%%%%%%%%%%%%%%%%%%%%%%%%%%%%%%%%%%%%
{\bfseries Non-local Operation}
Non-local operation define a generic non-local operation in deep neural networks as:
\begin{equation}
\boldsymbol{y}_i = \sum_{j\in\mathbb{D}^3}f(\boldsymbol{x}_i,\boldsymbol{x}_j)g(\boldsymbol{x}_j).
\end{equation}
Here function $f$ representing the relation between position $i$ and $j$. Many visions of function $f$ such as $f(\boldsymbol{x}_i,\boldsymbol{x}_j) = \exp{[\theta(\boldsymbol{x}_i)^T\varphi(\boldsymbol{x}_j)]}$ are discussed, but performed almost the same.

As done in\cite{carreira2017quo}\cite{feichtenhofer2016spatiotemporal}, a 2D $k\times k$ kernel can be inflated as a 3D $t\times k\times k$ kernel that spans t frames, in our experiments we used 32 frames. So this kernel can be initialized from 2D models(pretrained on Imagenet), each of the t planes in the $t\times k\times k$ kernel is initizlized by pretrained  $k\times k$ weights, rescaled by 1/t. Each video we  sample 64 consecutive frames from the original full-length video and then dropping every other frame.
The non-local operation computes the response at a position as a weighted sum of the features at all positions with Embedding Gaussian. We used 5 non-local blocks added to i3d baseline. Table \ref{3} shows our non-local results on Kinetics and Moments in Times dataset.
\begin{table}[h]
   \centering
   \begin{tabular}{p{3.5cm}<{\centering}|p{2cm}<{\centering}|p{2cm}<{\centering}}
      \hline
      Models      & Kinetics & MIT   \\ \hline
      Res50 baseline      & 78.63              & 30.83               \\ \hline
      Res50 non-local      & 80.80              & 32.96                  \\ \hline
      Res101 baseline     & 79.58               & 31.33                 \\ \hline
      Res101 non-local     & 81.96            & 33.69              \\ \hline
   \end{tabular}
   \caption{Performance of nonlocal NN on Kinetics and MIT dataset.}
   \label{3}
\end{table}

\section{Relation-driven Models}

We are interested in two questions. Firstly, $\emph{non-local}$ operations would be important for relation learning, but $\emph{global}$ operations may be unnecessary. If position $i$ is far away from $j$, then $f(\boldsymbol{x}_i,\boldsymbol{x}_j)\approx 0$. Second quesion is that an unsupervised function may not be able to learn relations.

{\bfseries Mask Non-local}
To answer the first question, we compared the performance of non-local opeations and mask non-local opeations:
\begin{equation}
\boldsymbol{y}_i = \sum_{j\in\mathbb{D}^3}\mathbb{I}_{\mathbb{D}_i}(j)f(\boldsymbol{x}_i,\boldsymbol{x}_j)g(\boldsymbol{x}_j).
\end{equation}
Here $\mathbb{D}_i$ is the $\delta -$ neighbourhood of $i=(t_i,h_i,w_i)$. Say:
\begin{equation}
\mathbb{D}_i = [t_i-\delta_t,t_i+\delta_t]\times[h_i-\delta_h,h_i+\delta_h]\times[w_i-\delta_w,w_i+\delta_w].
\end{equation}
$\mathbb{I}_{\mathbb{D}_i}(j)$ is the mask function. Say:
\begin{equation}
\mathbb{I}_A(x)=
\begin{cases}
1& \text{$x\in A$}\\
0& \text{$x\notin A$}
\end{cases}.
\end{equation}

Table \ref{123456} shows mask nonlocal's performance on Kinetics. $+\infty$ means non-local operation in the dimention. Note that the first setting is the non-local baseline.
\begin{table}[h]
   \centering
   \begin{tabular}{p{1.5cm}<{\centering}|p{1.5cm}<{\centering}|p{1.5cm}<{\centering}|c}
      \hline
      $\delta_t$    & $\delta_h$     & $\delta_w$     & top-1 acc\\\hline
      $+\infty$     &$+\infty$       &$+\infty$       &80.80\\\hline
      $+\infty$     &$\frac{3}{7}H$  &$\frac{3}{7}W$  &81.26\\\hline
      $+\infty$     &$\frac{3}{28}H$ &$\frac{3}{28}W$ &80.63\\\hline
      $\frac{1}{2}T$&$\frac{3}{7}H$  &$\frac{3}{7}W$  &81.65\\\hline
      $\frac{1}{2}T$&$\frac{3}{28}H$ &$\frac{3}{28}W$ &80.93\\\hline
   \end{tabular}
   \caption{Perfomance for different settings of $\delta$− neighbourhood.}
   \label{123456}
\end{table}
%%%%%%%%%%%%%%%%%%%%%%%%%%%%%%%%%%%%%%%%%%%
%%%%%%%%%%%%%%%%%%%%%%%%%%%%%%%%%%%%%%%%%%%

{\bfseries Learning Relations in Video}
Common convolution layers use invariant kernels for feature extraction at all positions in the feature map. It's limited for learning relations between different positions on the feature map. Nonlocal operations compute a feature-map-wise relation matrix to represent the kernel so that different positions get different but related feature extractions. The problem is that an unsupervised function may not be able for relations learning. We proposed a new model to learn the relation patten.

The network contains a network-in-network with a $(2t_0+1)\times (2h_0+1)\times (2w_0+1)$ size receptive field. The network-in-network computes a $(2t_1+1)\times (2h_1+1)\times (2w_1+1)$-dimensional relation vector $r^{(i)}$ for any position $i=(t,h,w)$ at the feature map ($t_1<t_0<\frac{1}{2}T$, $h_1<h_0<\frac{1}{2}H$ and $w_1<w_0<\frac{1}{2}W$ ).

\begin{figure}[h]
\centering\includegraphics[width=8cm]{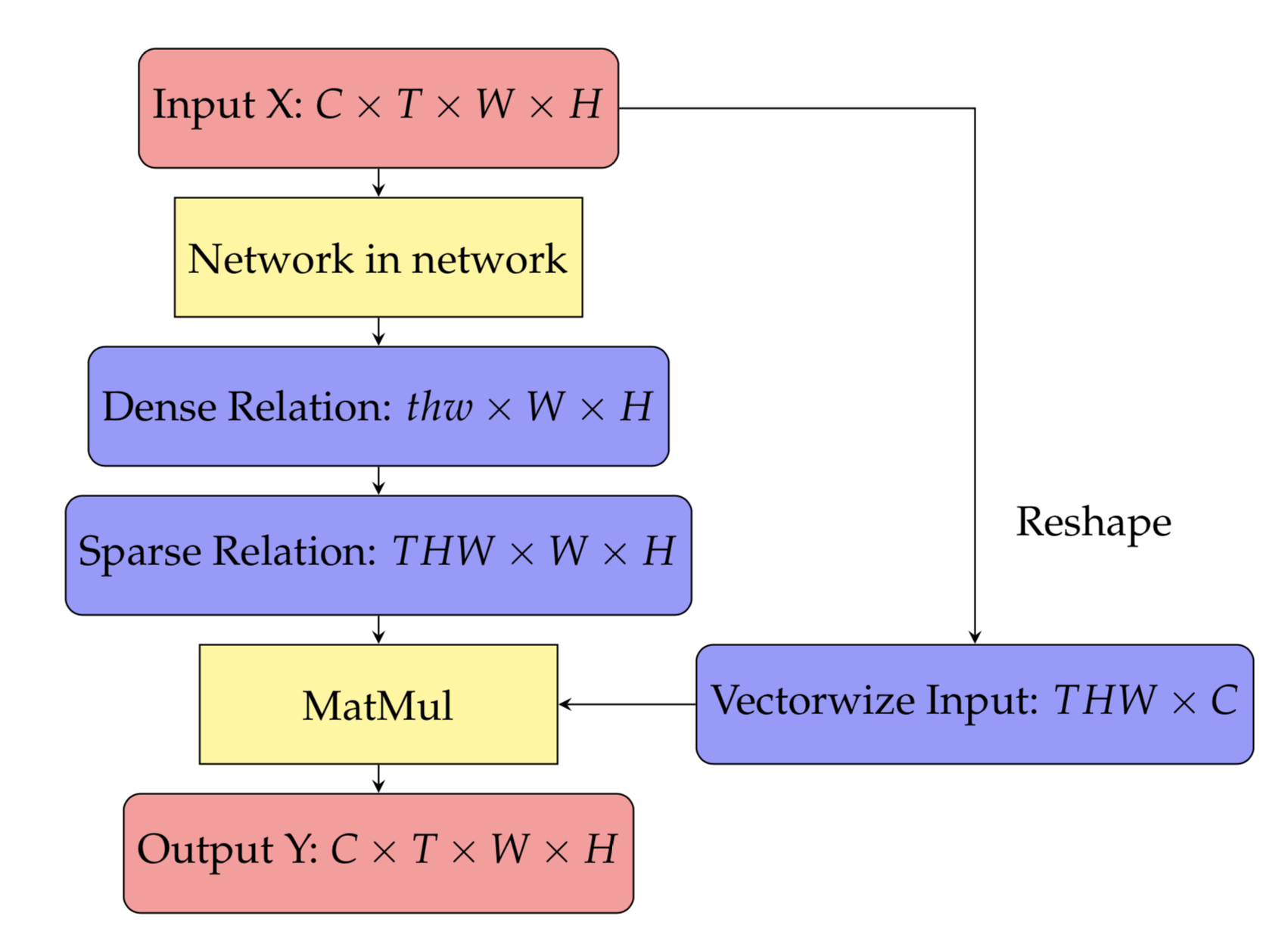}
\caption{Our network structure for learning relations}\label{666}
\end{figure}

The learnable relation vector $r^{(i)}$ represent the relation between position $i$ and its neighbourhood
\begin{equation}
\begin{split}
\mathbb{D}_i &= [t_i\pm t_1]\times[h_i\pm h_1]\times[w_i\pm w_1],\\
\boldsymbol{y}_i &= \sum_{j\in\mathbb{D}_i}r^{(i)}_jg(\boldsymbol{x}_j).
\end{split}
\end{equation}
Here $r^{(i)}_j$ is the $j^{th}$  element of $r^{(i)}$. Figure \ref{666} shows our network structure. Note that, by using mask non-local's initialization, our network can get better results than what table \ref{123456} shows. But due to time limit and training from scratch, we haven't finished the experiments.
\section{Acknowledgements}
Yingbin Zheng and Hao Ye from Shanghai Advanced Research Institute, Chinese Academy of Sciences also make contributions to this paper.

\renewcommand\refname{Reference}
\bibliographystyle{plain}
\bibliography{activitynet2018_qiniu_submit.bib}
\end{document}